# Robust Principal Component Analysis for Background Estimation of Particle Image Velocimetry Data

Ahmadreza Baghaie
Department of Electrical and Computer Engineering
New York Institute of Technology
Long Island, NY, USA

*Abstract*—Particle Image Velocimetry (PIV) data processing procedures are adversely affected by light reflections and backgrounds as well as defects in the models and sticky particles that occlude the inner walls of the boundaries. In this paper, a novel approach is proposed for decomposition of the PIV data into background/foreground components, greatly reducing the effects of such artifacts. This is achieved by utilizing Robust Principal Component Analysis (RPCA) applied to the data matrix, generated by aggregating the vectorized PIV frames. It is assumed that the data matrix can be decomposed into two statistically different components, a low-rank component depicting the still background and a sparse component representing the moving particles within the imaged geometry. Formulating the assumptions as an optimization problem, Augmented Lagrange Multiplier (ALM) method is used for decomposing the data matrix into the low-rank and sparse components. Experiments and comparisons with the state-of-the-art using several PIV image sequences reveal the superiority of the proposed approach for background removal of PIV data.

I. INTRODUCTION

Particle Image Velocimetry (PIV) as a standard optical imaging technique for flow velocity measurements has gained extensive interest. In this technique, $\mu$m-scale tracer particles are added to the flow, illuminated by a directed plane laser light and the scattered laser light is then recorded over the course of the imaging. This is followed by computational procedures to infer the displacements/velocities of the tracer particles. The processing is done by sectioning the PIV frames into small subregions, followed by cross-correlation-based particle tracking approaches, with the assumption of having homogeneous movements within each sub-region [1-3]. While single plane PIV imaging has its merits, in many real world applications complex three-dimensional (3D) flow patterns are observed where 2D PIV image acquisition/analysis faces great challenges. This has resulted in introduction of new PIV imaging techniques aiming at providing a more truthful capture of the 3D flow patterns. Among them, dual-/multi-plane PIV [4, 5], tomographic PIV [6, 7] and stereo PIV [8, 9] can be mentioned.

Various sources of artifacts and errors often compromise the subsequent processing of the PIV data. Camera dark noise, light scattering between particles, laser light reflections, and differences between refractive indexes of the geometry and working fluid are the major sources of errors and artifacts. Moreover, artifacts that block the cameras' view of the moving particles, such as defects of the model, stuck particles and stents, should be properly removed using image processing methodologies to reduce the bias when estimating the velocity fields [10, 13].

Methods of normalization and levelization are often used for pre-processing of the PIV data. In normalization methods, the aim is to increase the dynamic range of the PIV data to achieve higher contrast between the particles and background [14-16]. While such techniques are useful in enhancing the separability between the particles and their surroundings, they offer limited performance in completely removing the effects of background. Especially in the cases where the brightness of the background is in the same order as the particles. On the other hand, levelization techniques are generally used for background removal by estimating a reference intensity map of the background which is subtracted from each PIV frame in the sequence [17]. Background estimation in such methods is achieved by various means, from average/minimum intensity to sliding median/low-pass filters [9, 18].

Recently Mendez et al. [10] proposed a Proper Orthogonal Decomposition (POD) based approach for background removal of PIV data. In their work, the differences in spatial and temporal coherence of the background and particles are considered as basis for decomposition of the PIV image sequences into background and moving particles components. While the approach performs well in the provided test cases, there are some shortcomings that limit the applicability of the method. The formulation of the POD analysis is based on the classic Principal Component Analysis (PCA). In this method, the dimentionality reduction of the noisy data is considered by means of 2-norm minimization. Even though such formulation is useful in noise and dimentionality reduction of the data when it is degraded by noise and corruptions of low magnitude, it is highly sensitive to presence of grossly corrupted observations in the data [19]. In such cases, estimation of the underlying low dimensional manifold faces problems. These limitations result in erroneous decomposition of the PIV data into background/particles components which can compromise the subsequent analysis procedures. The effects can be seen in residual moving artifacts in the background components which come as the expense of the signal/intensity loss of the particles.

In this work, application of Robust Principal Component Analysis (RPCA) is investigated for background removal of PIV data. This is to address the limitations of the POD-based approach in decomposition of the PIV data into background/particles components. Here, we aim to encounter the problem from a purely image processing perspective and a stricter view of the decomposition process is considered that works based on the assumption of having highly low-rank background and sparse components in the aggregated data matrix. This assumption leads to better capability in decomposition of the background and particles components in PIV data.

The rest of the paper is organized as follows. In Section II, at first, an overview is given for Robust Principal Component Analysis (RPCA). Then, the problem of PIV data decomposition is explained and formulated in the context of RPCA and solved using an Augmented Lagrangian Multiplier (ALM) framework. Section III provides results with comparisons and discussions with the state-of-the-art. Section IV concludes the paper.

## II. METHODS

### A. Robust Principal Component Analysis

Principal Component Analysis (PCA) has been a very popular tool for data processing and analysis with many applications in various fields of science and engineering [21-23]. Given a set of high-dimensional data, the main assumption in PCA is that the data lie near a much lower-dimensional linear subspace. Therefore, the process of the low-dimensional extraction involves computing the eigenvalues/eigenvectors of the data using Singular Value Decomposition (SVD) and then projecting the data onto the first few principal left singular vectors. While this leads to satisfying results in cases of having input data corrupted by independent and identically distributed (i.i.d) Gaussian noises with small magnitudes, larger corruptions with grossly large magnitudes can cause the PCA estimates to be erroneous. This has lead the researchers to devise more *robust* PCA algorithms to deal with such anomalies [24].

Given a large $m \times n$ data matrix $D \in \mathbb{R}^{m \times n}$, each column representing an $m$-dimensional data point, Robust Principal Component Analysis (RPCA) tries to find a decomposition of the data matrix in the form [19]:

$$D = L + S \qquad (1)$$

where $L$ has low rank and $S$ is sparse. In another word, similar to that of PCA, the RPCA is assuming that the date points lie on some low-dimensional sub-space, while degraded by sparse noise/corruptions. This problem appears in different areas of computer vision and image processing, such as video surveillance for detection of moving objects in a constant background, face recognition for removing of shadows and specularities or even biomedical image processing [25, 26].

In an $l_2$ norm sense, solving Eq. (1) is equivalent to the classic PCA, which tries to minimize the power of the error between the approximated sub-space and the data points. But the classic PCA is very sensitive to outliers, therefore even with a small portion of data points affected by noise or large magnitudes of artifacts due to the measurements, it performs poorly. This is inherent to the $l_2$ formulation of the PCA as, even though leads to a unique analytical solution, the effects of grossly large corruptions are amplified because of the quadratic formulation. RPCA, on the other hand, is formulated as:

$$\min rank(L) + \lambda ||S||_0 \quad s.t. \quad D = L + S \qquad (2)$$

where $||.||_0$ is the $l_0$ norm measuring the sparsity of $S$ and $\lambda$ is a positive weighting parameter balancing the two terms. However, since this is an NP-hard problem [19], in practice, a convex relaxed version is considered as follows:

$$\min ||L||_* + \lambda ||S||_0 \quad s.t. \quad D = L + S \qquad (3)$$

where $||.||_*$ represents the nuclear norm (sum of singular values) of a matrix and $||.||_1$ represents the $l_1$ norm (sum of the absolute values of the entries) of a matrix. The solution to this convex optimization problem will result in recovery of the sparse and low-rank components of the data matrix $D$, with substantial reduction of the adverse effects of grossly large corruptions.

### B. PIV Data Foreground/Background Decomposition Using RPCA

An identical analogy can be made when aiming to estimate background in a set of PIV images. The background is always present in the sequence, with possible illumination variations, while the moving particles can be considered as the sparsely distributed components within the images. To achieve the decomposition using RPCA, vectorization and aggregation of the data is the first step. This results in a $m \times n$ data matrix $D$, where $m$ is the total number of pixels in each frame and $n$ is the number of frames in the dataset.

We aim to distinguish between the foreground moving particles within the geometry and the still background. For this, the optimization problem shown in Eq. (3) should be solved. This is achieved by employing Augmented Lagrange Multiplier (ALM) method. The general method of ALM for solving the constrained problem of

$$\min f(X) \quad s.t. \quad h(X) = 0 \qquad (4)$$

for functions $f: \mathbb{R}^n \to \mathbb{R}$ and $h: \mathbb{R}^n \to \mathbb{R}^m$ can be formulated by defining the augmented Lagrangian function $\mathcal{L}$ [27]:

$$\mathcal{L}(X, Y, \mu) = f(X) + <Y, h(X)> + \frac{\mu}{2} ||h(X)||_F^2 \qquad (5)$$

with $<.,.>$ representing the inner product, $||.||_F$ the Frobenius norm and $\mu$ a positive scalar. For RPCA, the problem can be formulated similarly by identifying the various components as:

$$\begin{aligned} X &= (L, S), \\ f(X) &= ||L||_* + \lambda ||S||_1, \\ h(X) &= D - L - S \end{aligned} \qquad (6)$$

Given this, the Lagrangian function can be written as:

$$\begin{aligned} \mathcal{L}(L, S, Y, \mu) &= ||L||_* + \lambda ||S||_1 + \\ &\quad <Y, D - L - S> + \frac{\mu}{2} ||D - L - S||_F^2 \end{aligned} \qquad (7)$$

To iteratively solve Eq. (7) for $L$ and $S$ at each iteration $k$, at first:

$$L_{k+1} = \arg \min_A \mathcal{L}(A_K, S_k, Y_k, \mu_k) = U\Gamma_{\mu_k^{-1}}[S]V^T \quad (8)$$

where $(U, S, V) = SVD(D - S_k + \mu_k^{-1}Y_k)$ and $\Gamma_\epsilon$ is a shrinkage operator eliminating the eigenvalues within $[-\epsilon, \epsilon]$. The next step is to update $S_k$ which is done by:

$$S_{k+1} = \arg \min_S \mathcal{L}(L_k, S_k, Y_k, \mu_k) = \Gamma_{\lambda\mu_k^{-1}}[D - L_{k+1} + \mu_k^{-1}Y_k] \quad (9)$$

These two main steps are followed by updating $Y_{k+1} = Y_k + \mu_k(D - L_{k+1} - S_{k+1})$ and $\mu_{k+1} \leftarrow 1.5\,\mu_k$. After convergence, the data matrix $D$ is decomposed into the low-rank $L$ and sparse $S$ components as expected. The two resulted matrices are reshaped into the original PIV data dimensions. As per Candes et al. [19], $\lambda = 1\sqrt{N}$, with $N$ being the total number of pixels in each frame. For more elaboration on the implementation aspects of the procedures discussed here as well as theories on the convergence of the methodology, the avid reader is referred to [28].

### III. RESULTS AND DISCUSSIONS

To assess the performance of the methods in background estimation, two sets of data are used here. For the first set, three synthetic frame sequences with known background and foreground are considered from the Background Models Challenge [29]. Each sequence contains 1499 color frames, each of size $480 \times 640$ pixels. For our experiments, only 100 consecutive frames from each sequence are considered. To assess the performance of the background/foreground decomposition methods, three metrics are used: Mean Squared Error (MSE), Peak Signal to Noise Ratio (PSNR) and Structural SIMilarity (SSIM). Assuming $F$ and $\hat{F}$ as the ground truth and estimated images respectively, the MSE can be defined as:

$$MSE = \frac{1}{N}\sum_{i=1}^{N}(f_i - \hat{f}_i)^2 \quad (10)$$

where $f_i$ and $\hat{f}_i$ are the $i$ th pixel of the ground truth and estimated images respectively, and $N$ is the total number of pixels. Having the MSE, PSNR can be defined as:

$$PSNR = 10\,log_{10}\frac{L^2}{MSE} \quad (11)$$

where $L$ is the dynamic range of pixel intensities in the images.

For the SSIM, three different components play significant roles: luminance, contrast ratio and structure. The simplified equation for SSIM can be written as [30]:

$$SSIM(F, \hat{F}) = \frac{(2\mu_F\mu_{\hat{F}} + C_1)(2\sigma_{F\hat{F}} + C_2)}{(\mu_F^2 + \mu_{\hat{F}}^2 + C_1)(\sigma_F^2 + \sigma_{\hat{F}}^2 + C_2)} \quad (12)$$

where $\mu_F$ and $\mu_{\hat{F}}$ are the averages and $\sigma_F$ and $\sigma_{\hat{F}}$ are the variances of the ground truth and the estimated image, respectively while the $\sigma_{F\hat{F}}$ is the covariance value. $C_1$ and $C_2$ are constants defined as $C_1 = (0.01 \times L)^2$ and $C_2 = (0.03 \times L)^2$.

Table I provides results of the quantitative comparison between the performance of the three decomposition methods: *min-removal*, *POD-based* and the proposed *RPCA-based* approach. For each set, the quantitative metrics of both the background and the foreground estimates are provided and the best performing approach is highlighted in bold. The strict formulation of the *RPCA-based* approach yields the most accurate estimation of the foreground/background components of the sequences, and the *min-removal* approach's performance is the worst as expected. While the performance of the *POD-based* method is superior to the *min-removal*, it still lacks in proper decomposition of the foreground/background components. This can be better seen in Fig. 1. Fig. 1 (a) shows a sample frame from the sequence while (b) and (c) show the ground truth background and foreground images respectively. In (d) the results of the *min-removal* approach are depicted, the background on the left and the foreground on the right. The same is done for the *POD-based* and *RPCA-based* approaches in (e) and (f) respectively. The *min-removal* approach considers the minimum intensity value of each pixel over time as the background and subtracts it in each pixel location. The effect of such assumption can be seen in (d) as the decomposition is suboptimal. On the other hand, the performance of the *POD-based* approach is superior to the *min-removal* since it aims to decompose the sequence by minimizing the $L_2$ of the estimated background and the input sequence. Even though such minimization is proven to provide satisfactory results when the input data is corrupted by i.i.d Gaussian noises with small magnitudes, in cases with larger corruptions with grossly large magnitudes such as the ones presented here, the estimations are erroneous. This can be seen in (e) where the decomposition is not done perfectly and traces of both background and foreground components are still visible in the other component's estimate. The *RPCA-based* approach on the other hand, performs almost perfectly and can decompose the input sequence into foreground/background components without any visible errors, as evidenced by Table I and Fig. 1.

TABLE I PERFORMANCE COMPARISON OF MIN-REMOVAL, POD-BASED AND THE PROPOSED METHOD FOR BACKGROUND AND FOREGROUND ESTIMATION OF THE THREE SYNTHETIC SEQUENCES FROM THE BACKGROUND MODELS CHALLENGE [29].

|  |  | Set 1 | | Set 2 | | Set 3 | |
|---|---|---|---|---|---|---|---|
|  |  | ***Background*** | ***Foreground*** | ***Background*** | ***Foreground*** | ***Background*** | ***Foreground*** |
| MSE | *Min-removal* | 381.45 | 352.32 | 312.64 | 277.14 | 278.77 | 230.87 |
|  | *POD-based* | 27.48 | 8.61 | 26.23 | 3.62 | 20.99 | 2.66 |
|  | *Proposed* | **0.12** | **0.02** | **0.16** | **0.07** | **3.70** | **1.59** |
| PSNR | *Min-removal* | 22.31 | 22.66 | 23.18 | 23.70 | 23.67 | 24.50 |
|  | *POD-based* | 35.49 | 45.98 | 34.49 | 43.82 | 35.37 | 44.63 |
|  | *Proposed* | **57.14** | **64.78** | **55.97** | **60.30** | **42.44** | **46.10** |
| SSIM | *Min-removal* | 0.9225 | 0.8833 | 0.9443 | 0.9225 | 0.9159 | 0.3227 |
|  | *POD-based* | 0.9897 | 0.9708 | 0.9898 | 0.9775 | 0.9431 | 0.9612 |
|  | *Proposed* | **0.9993** | **0.9990** | **0.9995** | **0.9989** | **0.9781** | **0.9701** |

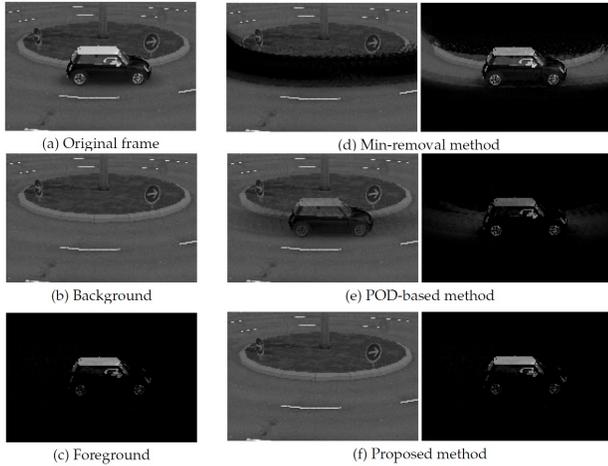

Fig. 1 Sample results from the Set 2 synthetic dataset. (a) original frame, (b) ground truth background, (c) ground truth foreground, (d) min-removal method, (e) POD-based method, (f) proposed method.

To assess the performance of the methods in background estimation of the PIV datasets, one set of synthetic and two sets of real PIV datasets, namely *Synthetic*, *Pipe* and *Stent* respectively, are utilized. The *Synthetic* set is generated by using an ideal PIV sequence of DNS simulation of a channel flow obtained from the John Hopkins Turbulence Database [31]. The ideal sequence is contaminated by various sources of background models and noise components, as implemented by Mendez et al. [10].

Fig. 2 shows the results of background removal of the synthetic sequence using three different methods: *min-removal*, *POD-based* and the proposed *RPCA-based* approach. In Fig. 2 (a) a sample frame from the sequence is shown while (b)-(d) show the results of the aforementioned methods. In each of (b)-(d), the left image is the background estimate and the right image is the particles components respectively. While the less computationally intensive *min-removal* approach is proven to be useful experimentally, its applicability is limited in cases where large variations are observed in the intensity levels of the background. In such cases, the estimation of the background cannot take into account such large variations and residual background artifacts are still present in the final result. In Fig. 2 (b) the contrast of the particles component is poor as a result of such residuals. As for the result of the *POD-based* approach the performance is better than the *min-removal* approach. However, careful examination of the result reveals that the decomposition of the initial frame into its background and particles components is not done perfectly as evidenced by the grainy appearance of the background. In other words, residual particles' traces can be seen in the background estimate of the frame. This is due to the $l_2$ formulation of the classic Principal Component Analysis (PCA) used in *POD-based* approach which relies on the 2-norm minimization of the decomposition problem. This makes it vulnerable to presence of grossly variable observations in the data [19]. In the current setup, such observations are the randomly distributed particles that compromise accurate decomposition of the initial frames. On the other hand, a more strict formulation of the decomposition problem as proposed in the proposed *RPCA-based* approach results in better performance in comparison to the *POD-based* approach. In Fig. (d) the background estimate does not contain residual particles' traces, therefore, providing a better decomposition and higher contrast for the particles component.

The same can be observed in the results of the *POD-based* method applied to the real PIV datasets, *Pipe* and *Stent*, as depicted in Fig. 3 and Fig. 4. In each, a sample initial frame is shown in sub-figure (a). The results of the decomposition into background and particles components using the *POD-based* and the proposed method are shown in (b) and (c). Close inspection of the results reveals the shortcoming of the *POD-based* approach in accurate estimation of the background as traces of particles can still be observed in the background. The problem is more prevalent in regions of slow-moving flow patterns, as in Fig. 4 for the *Stent* dataset. In these regions, *POD-based* approach cannot distinguish between the slow-moving particles and the still background properly and given that the particles component is generated by subtracting the background component from the initial frame, this results in signal loss in the particles component. In severe cases, this causes complete elimination of the moving particles and adversely affects the subsequent PIV processing. On the other hand, given the strict formulation of the *RPCA-based* approach, such artifacts are not observed, even in the regions of slow-moving flow.

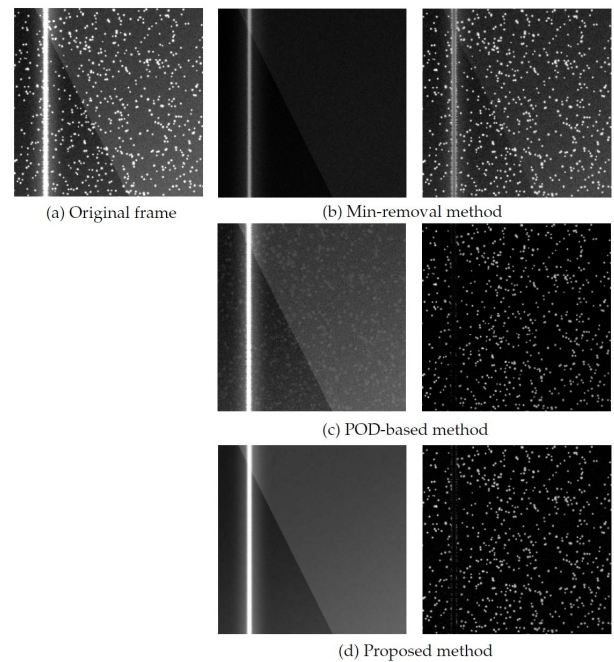

Fig. 2 Sample results from the Synthetic PIV dataset [10]. (a) sample original frame, (b) min-removal method, (c) POD-based method, (d) proposed method.

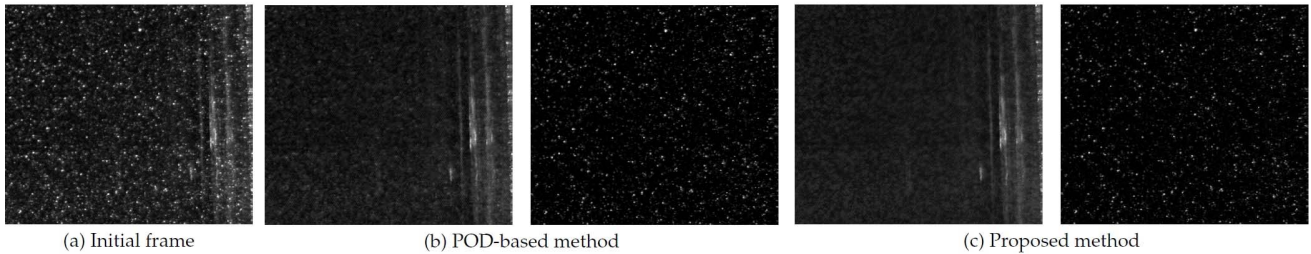

(a) Initial frame     (b) POD-based method     (c) Proposed method

Fig. 3 Sample results from the Pipe PIV dataset [10]. (a) sample initial frame, (b) POD-based method, (c) proposed method.

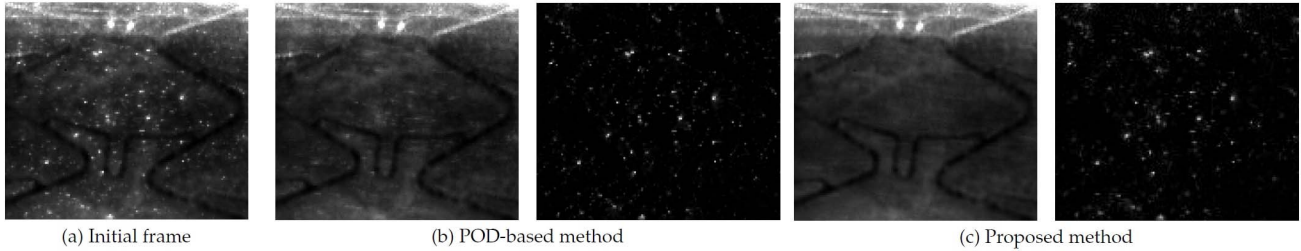

(a) Initial frame     (b) POD-based method     (c) Proposed method

Fig. 4 Sample results from the Stent PIV dataset [10]. (a) sample initial frame, (b) POD-based method, (c) proposed method.

## IV. Conclusion

In this work a novel approach for background removal of Particle Image Velocimetry (PIV) is proposed. The state-of-the-art is to adaptively truncate the POD bases in order to distinguish between the moving and stationary components in the PIV image sequence. While this can produce satisfactory results is some cases, it has been shown that the $l_2$ norm optimization of the Principal Component Analysis (PCA) employed in POD is not capable of full decomposition of the PIV data into background and particles components. The proposed Robust Principal Component Analysis (RPCA)-based approach is able to distinguish between the moving and stationary components more accurately. RPCA assumes that the aggregated data matrix, comprised by stacking vectorized versions of the frames in the PIV sequence, can be decomposed into two major components, a low-rank component depicting the background and a sparse component representing the sparse and randomly distributed moving particles. The decomposition is performed by formulating the problem as a convex optimization problem and the solution is computed by employing an augmented Lagrange multiplier optimization scheme. Experiments using both synthetic and real PIV image sequences show the superiority of the proposed approach in comparison to the state-of-the-art.

## Acknowledgment

The author would like to thank Dr. Bernhard Wieneke of LaVision GmbH for providing the *Pipe* and *Stent* PIV sequences. I also want to thank Mendez et al. [10] for providing the source code of their method and the *Synthetic* PIV sequence.